# 6 A Framework for Selection of Machine Learning Algorithms Based on Performance Metrices and Akaike Information Criteria in Healthcare, Telecommunication, and Marketing Sector

*A. K. Hamisu and K. Jasleen*

## CONTENTS

## 6.1 INTRODUCTION

> Machine intelligence is the last invention that humanity will ever need to make.
>
> *Nick Bostrom*

The growth of the internet has seen a profusion of data and a surge in technology for extracting information from big data for marketing strategy, adding value to products and services, and personalizing the consumer experience. Recently, there has been a remarkable increase in interest in the era of artificial intelligence (AI), ML, and deep learning (DL), as more individuals become aware of the breadth of new applications enabled by ML and DL methodologies. The applications of ML and DL range from home to hospital, domestic to enterprise, agriculture to military, and include all aspects of life. The main focus of this chapter is on applications of ML methodologies in three separate sub-domains: healthcare, marketing, and telecommunications. In the healthcare sector, two significant problems are considered for this research work. One is cardiovascular disease and another one is fetal health. The reason for choosing both these diseases is the rate at which they affect the people. Cardiovascular disease, also known as coronary ailment, is one of the most serious ailments in India and around the world. Heart disease is estimated to be the cause of 28.1% of deaths. It is also the leading cause of death, accounting for more than 17.6 million fatalities in 2016 across the world (Shan et al. 2017). As a result, accurate and early diagnosis and treatment of such diseases necessitates a system that can forecast with pinpoint accuracy and consistency. The second problem that is considered for this work is fetal health classification which includes classification of fetal as healthy or unhealthy. A total of three datasets (two cardiovascular datasets and one fetal health dataset) were used under healthcare sector. In this chapter, a framework for the selection of ML algorithm has been proposed. ML algorithm was selected based on dataset attributes, performance metrics, and AIC score. For experimentation purposes, ML algorithms were divided into eager, lazy, and hybrid learners. For the evaluation of the proposed framework, a total of eight datasets from three sectors (healthcare, telecommunication, and marketing) were selected for experimentation. This paper contributes in context of framework for recommendation of the best ML algorithm/model according to the input attributes. Model recommendation was based on performance evaluation parameters (accuracy, precision, and recall) as well as on model selection parameters (AIC).

The rest of the chapter is organized as follows. Section 6.2 presents related work carried out in proposed direction. Complete methodology followed for implementation of this work is presented in Section 6.3. Detailed results and analysis are presented in section 6.4 followed by concluding remarks in section 6.5.

## 6.2 MACHINE LEARNING APPLICATIONS

ML has potential applications in various domains and sectors. This section provides a brief glimpse of applications of ML in healthcare, telecommunication, marketing, and other sectors.

Goyal et al. (2021) introduced the concept of Internet of Health Things and discusses about potential challenges, advancement and benefits for IoT based healthcare and healthcare aided living. Pattnayak and Jena (2021) discussed and explained the need of ML for healthcare systems. Potential application of ML in healthcare and healthcare aided areas which includes from patient to doctor, from diagnosis to treatment, from surgery to decision support system were well elaborated. Panigrahi et al. (2021) developed an expert system-based clinical decision support system (CDSS) for prediction and diagnosis of hepatitis-B. This system comprises 59 rules and implementation is done using web-based Expert System Shell. Paramesha et al. (2021) discussed ML-based approach for sentiment analysis of narrated drug reviews and engineering in food technology which are indirectly related to the healthcare sector. Mohapatra et al. (2021) experimented with convolutional neural network (CNN) for early detection of skin cancer. They have also performed comparative analysis of MobileNet and ResNet50 CNN architectures for skin cancer classification task. Ramakrishnudu et al. (2021) proposed a system that predicts the overall health status of a person using ML techniques. Various parameters such as person's sleeping pattern, his/her physical activity, and his/her eating habits were used for predicting the overall health of the person. Panicker et al. (2021) proposed lightweight CNN model for classifying tuberculosis bacilli from non-bacilli objects. The performance of the proposed model in terms of accuracy is close to the existing ML models Panicker et al. 2021). Islam et al. discussed the use of DL techniques for autonomous disease diagnosis from symptoms. They proposed a graph convolution network (GCN) as a disease–symptom network to link the disease and symptoms. GCN-based deep neural network determines the most probable diseases associated with the given symptoms with 98% accuracy (Islam et al. 2021). Khamparia et al. (2020) proposed transfer learning based novel DL internet of health and things driven method for skin cancer classification. The proposed method performed well as compared to earlier reported techniques. Güldoğan et al. (2021) proposed a transfer learning-based technique for the detection and classification of breast cancer (benign or malignant) based on the ultrasound images. Performance metrics such as accuracy, sensitivity, and specificity with 95% confidence intervals were 0.974 (0.923–1.0), 0.957 (0.781–0.999), and 1 (0.782–1.0), respectively (Güldoğan et al. 2021). Said et al. (2021) proposed a new transfer learning-based approach for the classification of breast cancer in histopathological images. Block wise fine tuning strategy has been employed to handle CNN RESNET-18 (Said et al. 2021). Yang et al. (2021) explored the potential of DL models in the identification of lung cancer subtypes and cancer mimics from whole slide images. Irene et al. (2021) elaborated the ethics of ML in healthcare through the lens of social justice. Recent developments, challenges, and solutions to address those challenges were discussed in detail. Danton et al. (2020) proposed a systematic approach to identify the ethics in ML-based healthcare applications. Elements such as conceptual model, development, implementation, and evaluation were considered while framing the approach (Danton et al. 2020). Muhammad et al. (2020) discussed the challenges, requirements, and opportunities in the area of fairness in healthcare AI and the various nuances associated with it. Liu et al. (2020) proposed DL approaches for automatic diagnosis of Alzheimer's disease (AD) and its prodromal stage, that is, mild cognitive impairment (MCI). Baskar et al. (2020) proposed a framework for wearable sensors (WS) so that it can be applicable as a part of smart healthcare tracking applications. Andre et al. (2019) discussed the application of computer vision, natural language processing in the context of medical domain. Siddique and Chow (2021) discussed the application of ML/AI in healthcare communication. This work includes chatbots for the COVID-19 health education, cancer therapy, and medical imaging. The challenges, issues, and problems for the implementation of ML- and DL-based applications in healthcare and healthcare-aided sector have been discussed (Riccardo et al. 2018). Mateen et al. (2020) presented a framework for improving the accuracy of ML algorithms in healthcare by incorporating reporting guidelines such as SPIRIT-AI and CONSORT-AI in clinical and health science in ML approaches. Ferdous et al. (2020) presented a review on ML when applied to prediction of different diseases. The contribution of ML in healthcare is discussed with aim to provide the best suitable ML algorithm (Ferdous et al. 2020). Utsav et al. (2019) presented a technique to use ML algorithms for predicting the probability of cardiac arrest based on various attributes. Zoabi et al. (2021) proposed an ML-based technique to predict whether an individual is infected with SARS-CoV-2 or not. The model takes different parameters such as age, gender, and presence of various COVID symptoms. Faizal and Sultan (2020) explored the application of AI and data analytics techniques for mobile health. These techniques can be used for providing valuable insights to users and accordingly resources can be planned for mobile health. AI-based models have been proposed for mobile health. Futoma et al. (2020) emphasized for clinical utility and generalizability of ML algorithms and answers the various questions (when, how, and why) on ML applicability for both clinicians and for patients. Wang et al. (2020) proposed an alternative COVID-19 diagnosis methodology based on COVID-19 radio graphical changes in computerized tomography (CT) images. They experimented with DL methods to extract the hidden features from CT scans and provide the diagnosis for COVID-19 (Wang et al. 2020). Song et al. (2020) proposed DeepPneumonia technique (as DL based COVID detection from CT scans) to identify patients with COVID-19. Punn et al. (2020) proposed ML- and DL-based model to analyze predictive behavior of COVID-19 using a dataset published on the Johns Hopkins dashboard.

Authors proposed a technique to predict the customer churn rate (who are likely to cancel the subscription). Various ML algorithms such as DT, Random Forest, and XGBoost have been experimented (Kavitha et al. 2020). Researchers presented analysis to leverage ML methods in marketing research. Comparison between ML methods with statistical methods was also presented. A unified conceptual framework for ML methods have been proposed in this work (Liye and Baohong 2020). Dev et al. (2016) used ML to predict heart disease.

In Galván et al.'s (2009) study, a lazy learning strategy was proposed for building classification learning models. In this work, authors compared the accuracy of SVM and KNN algorithms on student performance data sets. SVM performed well as compared to KNN with accuracy of 91.07% (Nuranisah et al. 2020). Thanh and Kappas (2017) examined and compared the performance of ML algorithm for land use/cover classification. The classification results showed a high overall accuracy of all the algorithms In this paper, authors experimented with ML algorithms on healthcare datasets (Raj and Sonia 2017). Zhenlong et al. (2017) explored the usefulness of ML algorithms for driver drowsiness detection. The results revealed that SVM performed well. In this paper, authors proposed the application of lazy learning techniques to Bayesian tree induction and presents the resulting lazy Bayesian rule learning algorithm, called Lbr (Zheng and Webb 2000). Solomon et al. (2014) presented evaluation of eager and lazy classification algorithms using UCI Bank Marketing data set. Results revealed that eager learners outperform the lazy learners with accuracy of 98%.

## 6.3 DESIGN AND IMPLEMENTATION OF FRAMEWORK FOR MODEL SELECTION

Proposed architecture is explained in Figure 6.1. Proposed system consists of various phases: data collection, data pre-processing, feature extraction, model building, and performance evaluation.

### 6.3.1 Phase 1: Input Analysis Phase

#### 6.3.1.1 Input Attributes

In this phase, attributes are input into the system. The selection of attributes entirely depends upon the problem for which the most suitable algorithm is to be identified.

#### 6.3.1.2 Attribute Analysis

In this sub-phase, input attributes are analyzed. Various kinds of analysis such as size of input attributes, type of input attributes, and nature of input attributes are performed for better understanding of data. Visualization technique was used to identify the relationship between input attributes whether it is linear or non-linear, based on which set of ML algorithms were selected. For input attributes where a linear relationship exists among the attributes, algorithms like SVM can be selected for initial evaluation. If the size of input attributes is small, algorithm like Naïve Bayes can be preferable for initial evaluation. As nature of data concerns, it depends upon the output target variable type. In our work, target variable is categorical in nature.

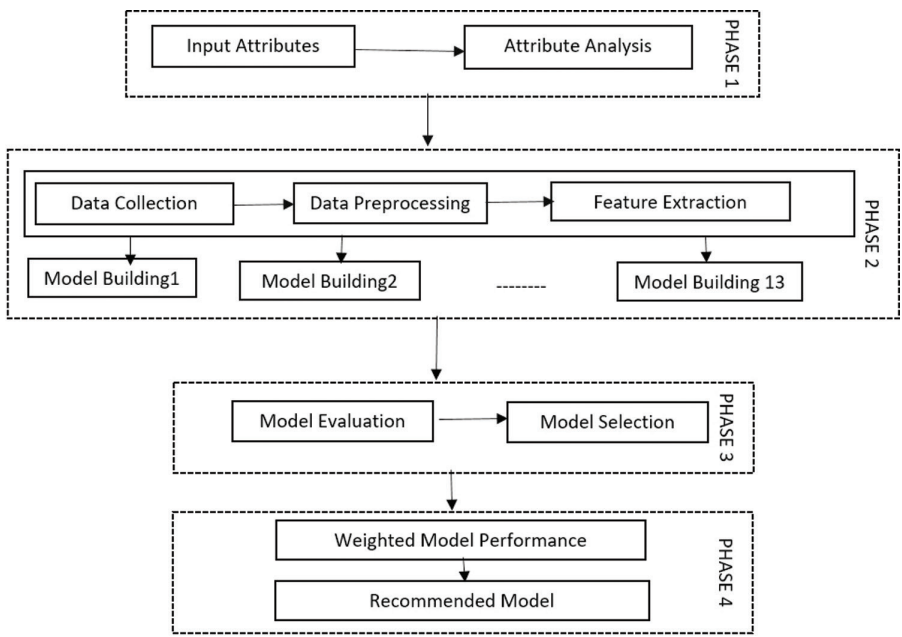

**FIGURE 6.1** Architecture of proposed system.

### 6.3.2 Phase 2: Model Building Phase

In this phase, various selected ML algorithms were trained and tested for datasets from each of the sectors. The ML algorithm training and testing process is described in the following sub-phases.

#### 6.3.2.1 Data collection

For this research work, three sectors (healthcare, telecommunication, and marketing) were selected. A total of eight datasets were collected in three sectors. Details of dataset by sector are presented in Table 6.1. The description includes the number of records, attributes, and class labels in each dataset.

#### 6.3.2.2 Data Pre-Processing

Raw data need to be pre-processed to organize them into the form which is good for training the ML algorithm (Han and Kamber 2001). In this work, raw data passes through various pre-processing stages such as label encoding and handling missing values with mean of that attribute.

#### 6.3.2.3 Features Extraction

To reduce the computational cost and time for building the model, subsets of features/attributes were selected. Multifactor dimensionality reduction methods were used for reducing the dimensionality of the data. Some attributes were not considered for model building based upon co-linearity matrices (used for finding relationship with indented class label).

**TABLE 6.1**
**Dataset Description**

| Sector | Dataset | Description |
|---|---|---|
| Marketing | (Avocado 2020) | 18,249 records; 13 attributes; 2 class labels |
| | (Bank in Marketing 2020) | 11,162 records; 17 attributes; 2 class labels |
| Telecommunication | (Telecom 2020) | 4000 records; 12 attributes; 2 class labels |
| | (Cell2cell train 2020) | 51047 records; 38 attributes; 2 class labels |
| | (Churn in Telecom 2020) | 3333 records; 21 attributes; 2 class labels |
| Healthcare | (Cardio-Vascular 2020) | 70,000 records; 13 attributes; 2 class labels |
| | (Fetal_Health 2020) | 2126 records; 22 attributes; 2 class labels |
| | (Health_heart 2020) | 1025 records; 14 attributes; 2 class labels |

#### 6.3.2.4 Model Building

In this research work, 13 machine algorithms were experimented. These ML algorithms were divided into the following categories: eager or lazy depending upon the learning procedure and third category is hybrid (Huang et al. 2014; Dev et al. 2016).

1. Eager learning: This category of ML algorithms includes DT, SVM, and Neural Network (NN).

   A DT is built using recursive partitioning-based approach. A tree-like structure is generated using input attributes and leaf nodes of those generated trees represent the class labels. In this research work, C4.5 version of DT was built using the gain ratio of attribute.

   $$Gain\ Rtio(X_i, D) = Information\ Gain\ (X_i, D) \Big/ Entropy\ (P_{X_i}(D)) \qquad (6.1)$$

   Where $Gain\ Ratio(X_i, D)$ is ratio of attribute $X_i$ with regard to Dataset D (Han and Kamber 2001).

   SVM is statistical ML algorithm which is based on the structural risk minimization principle (Han and Kamber 2001). Linear SVM tries to find maximal marginal hyperplane using the following equation:

   $$f(\vec{x}) = \begin{cases} 1 & \text{if } \vec{w}\cdot\vec{x} + b \geq 1 \\ -1 & \text{if } \vec{w}\cdot\vec{x} + b \leq -1 \end{cases} \qquad (6.2)$$

   Where $\vec{w}$ and $b$ parameters are identified from training data.

   NN is supervised ML algorithm which is based on backprogation where weights in hidden layer and output layer are updated according to error in estimation. For weights updation, the following equation is utilized.

   $$w_j^{k+1} = w_j^k + \lambda(y_i - \widehat{y_i^k})\,x_{ij} \qquad (6.3)$$

   Where k is iteration, $x_{ij}$ is input attribute value, $w_j^k$ is weight assigned in k[th] iteration, and $\lambda$ learning rate (Han and Kamber 2001).

2. Lazy learning: This category includes KNN algorithm and LNB algorithm.

   KNN algorithm is a distance-based ML algorithm which has application in classification as well as regression problems. In this research work, distance is calculated based on Euclidean distance measure. Distance between test point (x) and existing training point (y) is given by,

   $$Eucidean\ distance = \sqrt{\sum_{i=1}^{n}(x_i - y_i)^2} \qquad (6.4)$$

   For each dataset, hyper-parameter for KNN, that is, k, is tuned using elbow method.

**TABLE 6.2**

**Category-Wise Machine Learning Algorithms**

| S. No. | Category | Algorithm |
|---|---|---|
| 1 | Eager | Decision Tree (DT) |
|   |   | Support Vector Machine (SVM) |
|   |   | Neural Network (NN) |
| 2 | Lazy | K-nearest Neighbhour (KNN) |
|   |   | Lazy Naïve Bayes (LNB) |
| 3 | Hybrid | KNN+LNB |
|   |   | SVM+DT+NN |
|   |   | SVM+KNN |
|   |   | DT+KNN |
|   |   | NN+KNN |
|   |   | SVM+LNB |
|   |   | DT+LNB |
|   |   | NN+LNB |

3. Hybrid Learning: This category of ML algorithms was formed by combining different ML algorithms. Algorithms in this category are generated by stacking up the different ML algorithms from the eager and lazy categories. Count in eager, lazy, and hybrid ML categories is 3, 2, and 8 respectively. Table 6.2 provides the details about the categories of ML algorithms.

### 6.3.3 Phase 3: Model Evaluation Phase

#### 6.3.3.1 Model Analysis Module

In this phase, analysis of each ML model is carried out in terms of performance evaluation parameters and model selection parameters. Accuracy, precision, recall, F-measure, receiver operating characteristic (ROC) curve, and ROC area under curve (AUC) are used as performance metrics for evaluation. For model selection purposes, the AIC score was calculated for each algorithm (Akaike 1973).

### 6.3.4 phase 4: Model Recommendation Phase

Based on the attributes passed on in phase 1, this phase identifies the most suitable ML algorithm based on performance metrics and AIC score. Recommendation of ML algorithm is based on the weighted average of performance parameters and AIC score.

## 6.4 RESULT AND ANALYSIS

The purpose of this research is to find the best performing ML algorithm in each sector (telecommunication, health, and marketing). For this purpose, a total 104 experiments were performed where every dataset (8 in total) is experimented with 13 ML algorithms (as listed in the previous section). Selection of ML algorithm is carried out on the basis of performance parameters as well as AIC score. Implementation of this work has been carried out in Python.

### 6.4.1 Selection of Model based on Accuracy, Precision, Recall, and F-Measure

Tables 6.3–6.5 show the results of ML algorithms in each sector. For interpretation purposes, the average of each metric (accuracy, precision, recall, and f-measure) is obtained.

From Tables 6.3–6.5, it can be observed that the eager learner category of ML algorithms performed well as compared to lazy and hybrid learner. Overall average

**TABLE 6.3**
**Result Obtained with Marketing Sector**

| Learning Methods | Average Accuracy | Average Precision | Average Recall | Average F-measure |
|---|---|---|---|---|
| Eager learner | 94 | 0.92 | 0.99 | 0.95 |
| Lazy learner | 91 | 0.86 | 0.74 | 0.78 |
| Hybrid learner | 92 | 0.88 | 0.93 | 0.93 |

**TABLE 6.4**
**Result Obtained with Healthcare Sector**

| Learning Methods | Average Accuracy | Average Precision | Average Recall | Average F-measure |
|---|---|---|---|---|
| Eager learner | 90 | 0.88 | 0.83 | 0.88 |
| Lazy learner | 85 | 0.86 | 0.88 | 0.87 |
| Hybrid learner | 76 | 0.78 | 0.77 | 0.79 |

**TABLE 6.5**
**Result Obtained with Telecommunication Sector**

| Learning Methods | Average Accuracy | Average Precision | Average Recall | Average F-measure |
|---|---|---|---|---|
| Eager learner | 90 | 0.89 | 0.99 | 0.92 |
| Lazy learner | 86 | 0.90 | 0.84 | 0.86 |
| Hybrid learner | 85 | 0.78 | 0.87 | 0.88 |

### TABLE 6.6
**Average Accuracy-Based Comparison of Machine Learning Algorithms**

| Learning Methods | Marketing Dataset Average Accuracy | Telecommunication Dataset Average Accuracy | Healthcare Dataset Average Accuracy |
| --- | --- | --- | --- |
| Eager learner | 94 | 90 | 90 |
| Lazy learner | 91 | 86 | 85 |
| Hybrid learner | 92 | 85 | 76 |

### TABLE 6.7
**Average Precision-Based Comparison of Machine Learning Algorithms**

| Learning Methods | Marketing Dataset Average Precision | Telecommunication Dataset Average Precision | Healthcare Dataset Average Precision |
| --- | --- | --- | --- |
| Eager learner | 0.92 | 0.89 | 0.88 |
| Lazy learner | 0.86 | 0.90 | 0.86 |
| Hybrid learner | 0.88 | 0.78 | 0.78 |

accuracy of eager learners ranges from 90% to 94%. Accuracy- and precision-based comparative analysis is presented in Tables 6.6 and 6.7.

From Figures 6.2 and 6.3, it can be observed that the eager learner category of algorithms performed well for the healthcare sector based on accuracy and precision. For identification of the best ML algorithm in each sector, performance analysis of ML algorithms in the eager learner category is carried out. From Figures 6.2 and 6.3, it can be observed that the eager learner category of algorithms performed well for the healthcare sector based on accuracy and precision. For identification of the best ML algorithm in each sector, performance analysis of ML algorithms in the eager learner category is carried out. In the case of the healthcare dataset, SVM is proven to be the best ML algorithm, whereas in the case of the telecommunication and marketing dataset, DT comes out as the top performing one. NN was the worst performing algorithm in each sector. Furthermore, ROC curve (refer to Figures 6.4–6.6) and ROC-AUC score was analyzed for top performing algorithms. ROC-AUC score comes out to be 1.0 for all top performing ones.

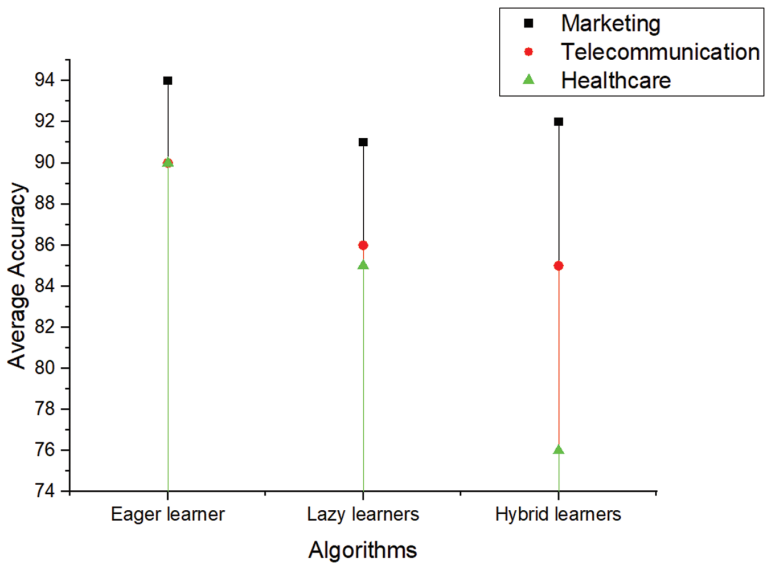

**FIGURE 6.2**　Comparison of algorithms based on accuracy.

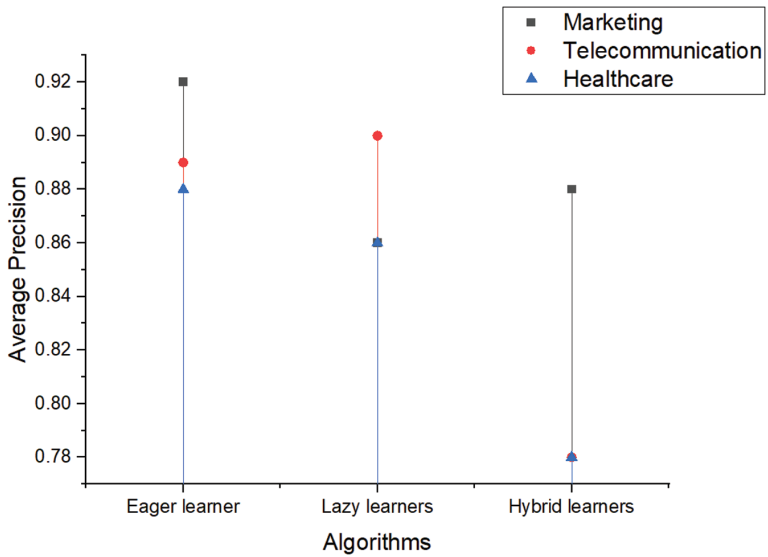

**FIGURE 6.3** Comparison of algorithms based on precision.

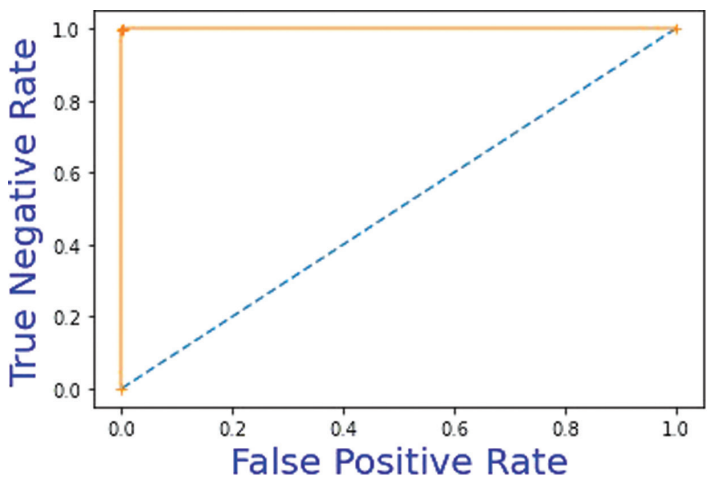

**FIGURE 6.4** ROC curve for DT algorithm in the marketing sector.

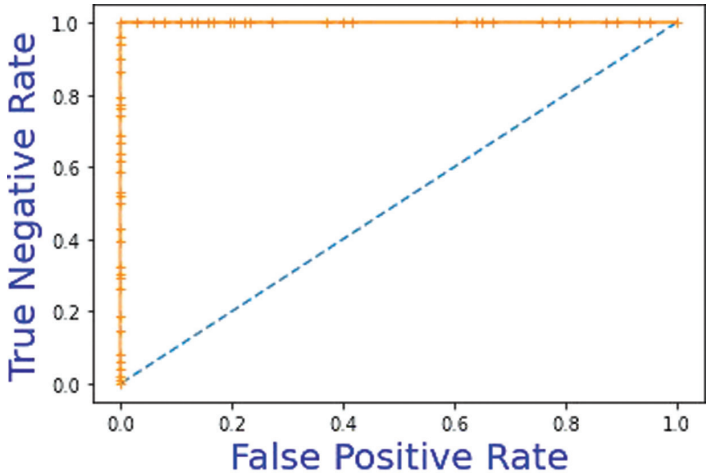

**FIGURE 6.5** ROC curve for SVM algorithm in the healthcare sector.

As performance of all the ML algorithms is on the same scale, further analyses is carried out using AIC.

### 6.4.2 Selection of Model Based on Akaike Information Criteria

In this section, algorithm performance is measured in terms of AIC. The best model is chosen with the help of probability framework of log-likelihood under maximum likelihood estimation. The AIC score can be calculated using:

$$AIC = 2*k - 2\ \log(L) \tag{6.5}$$

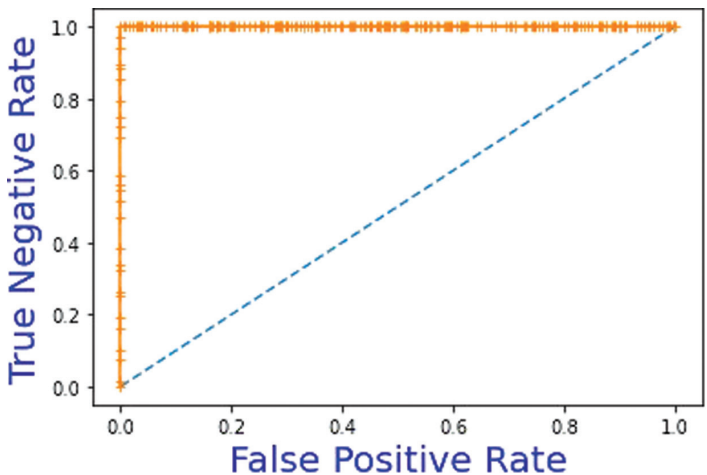

**FIGURE 6.6** ROC curve for DT algorithm in the telecommunication sector.

**TABLE 6.8**

**Results Based on Average AIC Score**

| Learning Methods | Marketing Dataset Average AIC | Telecommunication Dataset Average AIC | Healthcare Dataset Average AIC |
| --- | --- | --- | --- |
| Eager learner | 15.41 | 21.92 | 24.07 |
| Lazy learner | 16.35 | 19.91 | 21.76 |
| Hybrid learner | 17.38 | 22.19 | 24.96 |

where k indicates the number of independent variables used to build the model and L indicates maximum likelihood estimate of model (Akaike 1973). The best model is one which minimizes the information loss and has the minimum score for AIC.

From Table 6.8 it can be observed that in the marketing dataset, the lowest AIC score of 15.41 is reported by the eager learner category of ML algorithms, whereas for the telecommunication and healthcare datasets, the lowest AIC score is reported by the lazy learner category of ML algorithms with a score of 19.91 and 21.76, respectively.

From Figure 6.7 it can be observed that for the marketing sector, the lowest AIC score is reported by eager learners, whereas in the case of telecommunication and healthcare sectors, the lazy learner category reported the lowest AIC. To find the best suitable algorithm for each sector, comparative analysis has been

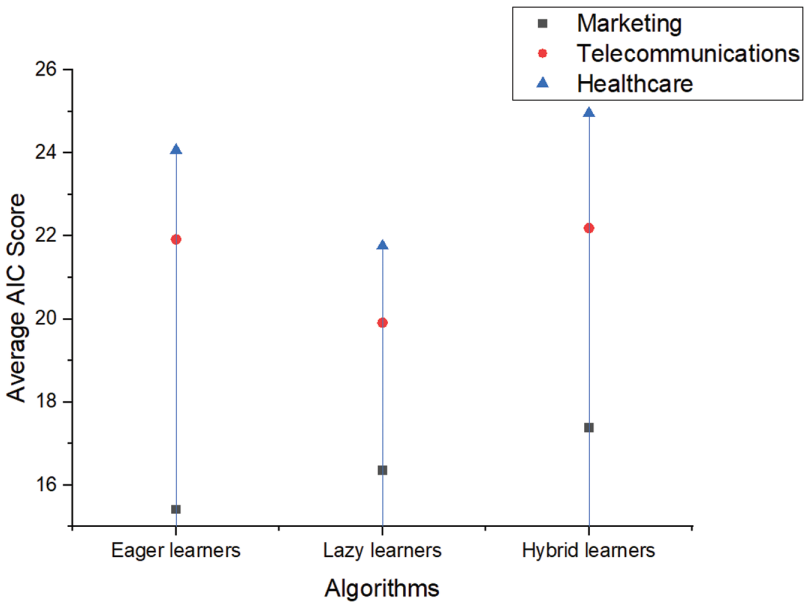

**FIGURE 6.7**   Comparison of algorithms based on AIC score.

**TABLE 6.9**

**Accuracy- v/s AIC Score-Based Comparative Analysis of Algorithms**

|  | Accuracy-Based Analysis | | AIC-Based Analysis | |
| --- | --- | --- | --- | --- |
|  | Category | Algorithm | Category | Algorithm |
| Marketing | Eager | DT | Eager | SVM |
| Telecommunication | Eager | DT | Lazy | KNN |
| Healthcare | Eager | SVM | Lazy | KNN |

carried out. Comparative analysis of algorithms based on accuracy and AIC score is presented in Table 6.9.

FIGURE 6.7 Three different ML model categories (lazy, eager, hybrid) for three different sectors are compared using AIC for model selection.

From Table 6.9 it can be observed that for the marketing dataset, the eager learner category is the better option as compared to lazy learners. Based on accuracy and AIC score, the most suitable ML algorithms are DT and SVM. On the basis of accuracy, eager learner is the best performing category of ML algorithms. DT and SVM are the most suitable algorithms for the telecommunication as well as healthcare sectors. On the basis of AIC score, lazy learner category of ML algorithms is the best option for the telecommunication and healthcare sectors. Out of all lazy learners, KNN performed well on telecommunication as well as on the healthcare sector.

## 6.5  CONCLUSION

In this research work, a framework for recommendation of ML algorithm has been formulated. The purpose was to find the most suitable ML algorithm for three different sectors. For experimentation purpose, ML algorithm were divided into three categories: eager, lazy, and hybrid learner. KNN, LNB, SVM, DT, NN, and the hybrid classifier using stacking were used on eight different datasets (from three different sectors: marketing, healthcare, and telecommunication). On the basis of accuracy, results revealed that eager learner ML algorithms are the best performing ones in all three sectors. Among eager learners, SVM is proven to be the top performing in healthcare with precision of 0.98. DT is the best suited for the telecommunication and marketing datasets with precision of 0.99 and 0.94, respectively. Whereas, on the basis of AIC score, SVM is the best suited for the marketing dataset, whereas KNN is the best suited for telecommunication and healthcare dataset.

Framework for Selection of ML Algorithms    **127**Avocado dataset accessed from https://www.kaggle.com/neuromusic/avocado-prices in Dec 2020.
Bank in Marketing dataset accessed from https://www.kaggle.com in Dec 2020.
Baskar, S., Mohamed, S. P., Kumar, R., Burhanuddin, M. A., Sampath, R. 2020. A dynamic and interoperable communication framework for controlling the operations of wearable sensors in smart healthcare applications. *Computer Communications*,149: 17–26. https://doi.org/10.1016/j.comcom.2019.10.004
Cardio_Vascular dataset accessed from https://www.kaggle.com in Dec 2020.
Cell2celltrain dataset accessed from https://www.kaggle.com in Dec 2020.
Churn in Telecom dataset accessed from https://www.kaggle.com in Dec 2020.
Danton, S. C., Michael, D. A., Chris, F. 2020. Identifying ethical considerations for machine learning healthcare applications. *The American Journal of Bioethics*, 20(11): 7–17. https://doi.org/10.1080/15265161.2020.1819469
Dev, S. K., Krishnapriya, S., Kalita, D. 2016. Prediction of heart disease using data mining techniques. *Indian Journal of Science and Technology*, 9(39): 1–5.
Faizal, K. Z., Sultan, R. A. 2020. Applications of artificial intelligence and big data analytics in m-health: A healthcare system perspective. *Journal of Healthcare Engineering*, 2020: 1–15. https://doi.org/10.1155/2020/8894694
Ferdous, M., Debnath, J., Chakraborty, N. R. 2020. Machine Learning Algorithms in Healthcare: A Literature Survey. *11th International Conference on Computing, Communication and Networking Technologies (ICCCNT)*, Kharagpur, India, 2020, pp. 1–6, https://doi.org/10.1109/ICCCNT49239.2020.9225642
Fetal_Health dataset accessed from https://www.kaggle.com in Dec 2020.
Futoma, J., Simons, M., Panch, T., Doshi-Velez, F., Celi, L. A. 2020. The myth of generalisability in clinical research and machine learning in health care. *The Lancet. Digital health*, 2(9): e489–e492. https://doi.org/10.1016/S2589-7500(20)30186-2
Galván, I. M., Valls, J. M., Lecomte, N., Isasi, P. 2009. A Lazy Approach for Machine Learning Algorithms. *Artificial Intelligence Applications and Innovations III. AIAI 2009. IFIP International Federation for Information Processing*, vol 296. Boston, MA: Springer. https://doi.org/10.1007/978-1-4419-0221-4_60
Goyal, S., Sharma, N., Bhushan, B., Shankar, A., Sagayam, M. 2021. *IoT Enabled Technology in Secured Healthcare: Applications, Challenges and Future Directions*. In Hassanien A.E., Khamparia A., Gupta D., Shankar K., Slowik A. (eds) *Cognitive Internet of Medical Things for Smart Healthcare. Studies in Systems, Decision and Control*, 311: 25–48. https://doi.org/10.1007/978-3-030-55833-8_2
Güldoğan, E., Ucuzal, H., Küçükakçalı, Z., Çolak, C. 2021. Transfer learning-based classification of breast cancer using ultrasound images. *Middle Black Sea Journal of Health Science*, 7(1): 74–80. https://doi.org/10.19127/mbsjohs.876667
Han, J., Kamber, M. 2001. *Data Mining: Concepts and Techniques*. San Diego, USA: Morgan Kaufmann.
Health_heart dataset accessed from https://www.kaggle.com in Dec 2020.
Huang, G., Song, S., Gupta, J. N. D., Wu, C. 2014. Semi-supervised and unsupervised extreme learning machines. *IEEE Transactions on Cybernetics*, 44(1): 2405–17. https://doi.org/10.1109/TCYB.2014.2307349
Irene, Y. C., Emma, P., Sherri, R., Shalmali, J., Kadija, F., Marzyeh, G. 2021. Ethical machine learning in healthcare. *Annual Review of Biomedical Data Science*, 4(1): 1–24. https://doi.org/10.1146/annurev-biodatasci-092820-114757
Islam, S. R., Sinha, R., Maity, S. P., Ray, A. K. 2021. Deep Learning on Symptoms in Disease Prediction. In: Mohanty, S. N., Nalinipriya, G., Jena, O. P. and Sarkar, A. (eds), *Machine Learning for Healthcare Applications,* Wiley Online Library, pp. 77–87. https://doi.org/10.1002/9781119792611.ch5
BK-TandF-JENA_9781032126876-211334-Chp06.indd   127    13/11/21   4:16 PM